\def\year{2020}\relax
\documentclass[letterpaper]{article} 
\usepackage{aaai20}  
\usepackage{times}  
\usepackage{helvet} 
\usepackage{courier}  
\usepackage[hyphens]{url}  
\usepackage{graphicx} 
\usepackage{graphicx}  
\usepackage{graphicx} 
\usepackage{amsmath}
\usepackage{mathptmx}  

\newcommand{\card}[1]{\lvert#1\rvert}			

\usepackage[utf8]{inputenc} 
\usepackage[T1]{fontenc}    

\title{Constrained deep neural network architecture search \\for IoT devices accounting hardware calibration\thanks{IBM, the IBM~logo, and ibm.com are trademarks or registered trademarks of International Business Machines Corporation in the United States, other countries, or both. Other product and service names might be trademarks of IBM or other companies.}
\thanks{Published as a conference paper at NeurIPS 2019}}

\author{Florian Scheidegger\textsuperscript{\rm 1,2}, Luca Benini\textsuperscript{\rm 1,3}, Costas Bekas\textsuperscript{\rm 2}, Cristiano Malossi\textsuperscript{\rm 2}\\
\textsuperscript{\rm 1} ETH Zürich, Rämistrasse 101, 8092 Zürich, Switzerland\\
\textsuperscript{\rm 2} IBM Research - Zürich, Säumerstrasse 4, 8803 Rüschlikon, Switzerland\\
\textsuperscript{\rm 3} Università di Bologna, Via Zamboni  33, 40126 Bologna, Italy\\
}

\begin{document}

\maketitle

\begin{abstract}
Deep neural networks achieve outstanding results in challenging image classification tasks. However, the design of network topologies is a complex task and the research community makes a constant effort in discovering top-accuracy topologies, either manually or employing expensive architecture searches. In this work, we propose a unique narrow-space architecture search that focuses on delivering low-cost and fast executing networks that respect strict memory and time requirements typical of Internet-of-Things (IoT) near-sensor computing platforms. Our approach provides solutions with classification latencies below $10ms$ running on a $\$35$ device with $1GB$ RAM and $5.6GFLOPS$ peak performance. The narrow-space search of floating-point models improves the accuracy on CIFAR10 of an established IoT model from $70.64\%$ to $74.87\%$ respecting the same memory constraints. We further improve the accuracy to $82.07\%$ by including 16-bit half types and we obtain the best accuracy of $83.45\%$ by extending the search with model optimized IEEE 754 reduced types. To the best of our knowledge, we are the first that empirically demonstrate on over 3000 trained models that running with reduced precision pushes the Pareto optimal front by a wide margin. Under a given memory constraint, accuracy is improved by over $7\%$ points for half and over $1\%$ points further for running with the best model individual format.
\end{abstract}

\section{Introduction}
\label{sec:intro}
With an increasing number of published methods, data, models, new available deep learning frameworks, and hype of special purpose hardware accelerators that become more commercially available, the design of an economical viable artificial intelligence system becomes a formidable challenge.
%
The availability of large scale datasets with known ground truth \cite{deng2009imagenet,data_gtsrb,data_cifar10_100} and widespread commercial availability of increased computational performance, usually achieved with graphics processing units (GPUs), enables the current growth of deep learning and explains the large interest and the emergence of new businesses.
Smart homes \cite{li2019smartHome}, smart grids \cite{fenza2019smartGrid} and smart cities \cite{gaber2019IOT_smartcities} trigger a natural demand for the Internet of Things (IoT), which are products designed around low cost, low energy consumption and fast reaction times due to the inherent constraints given by the final application that typically demand for autonomy with long battery lifetimes or fast real-time operation.
Experts estimate a number of around 30 billion IoT devices by 2020 \cite{fewer50BIotdevices} many of which serve applications that profit from artificial intelligence deployment.

In this context, we propose an automatic way to design deep learning models satisfying user-given constraints that are specially tailored to match typical IoT requirements, such as inference latency bounds. 
Additionally, our approach is designed in a modular manner that allows future adaptations and specializations for novel network topology extensions to different IoT devices and reduced precision arithmetic. 
In summary, our main contributions are the  following: 

\begin{itemize}
	\item We propose an end-to-end approach to synthesize models that satisfy IoT application and HW constraints.
	\item We propose a narrow-space architecture search algorithm to leverage knowledge from large reference models to generate a family of small and efficient models.
	\item We evaluate reduced precision formats for over 3000 models.
	\item We isolate IoT device characteristics and demonstrate how our concepts operate with analytical network properties and map to final platform specific metrics. 
\end{itemize}

The remainder of the paper is organized as follows. Section~\ref{sec:relatedwork} describes the related work, Section~\ref{sec:coredesign} introduces the core design procedures, Section~\ref{sec:cognitive} details and merges a full synthesis workflow, Section~\ref{sec:results} states and discusses the obtained results, and Section~\ref{sec:conclusion} concludes all findings.

\begin{figure}[t]
	\centering
	\includegraphics[width=0.9\columnwidth]{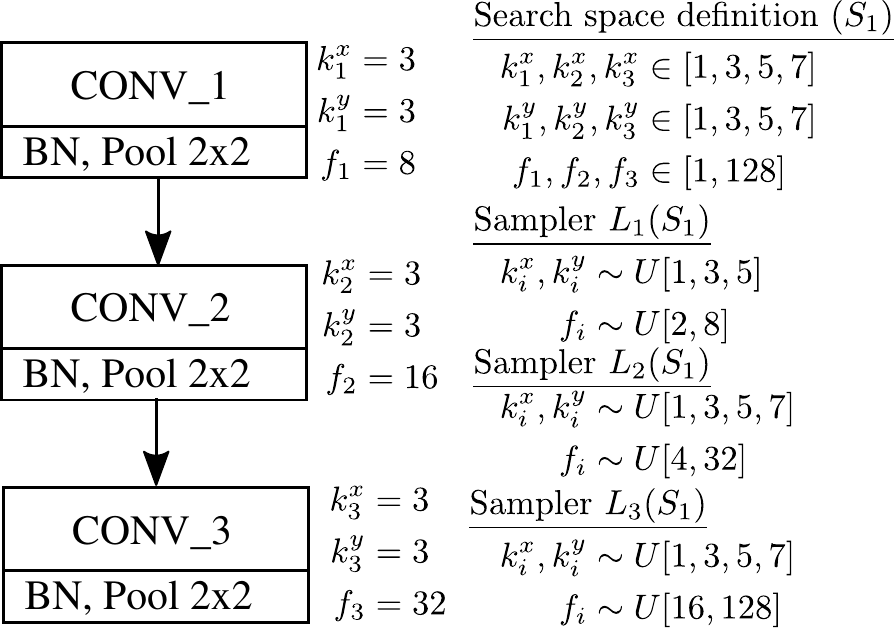}
	\caption{Simple three layer architecture with default configuration of search space with restricted sampling laws.}
	\label{fig:fig_arch_serch_I}
\end{figure}

\begin{figure}[t]
	\centering
	\includegraphics[width=0.9\columnwidth]{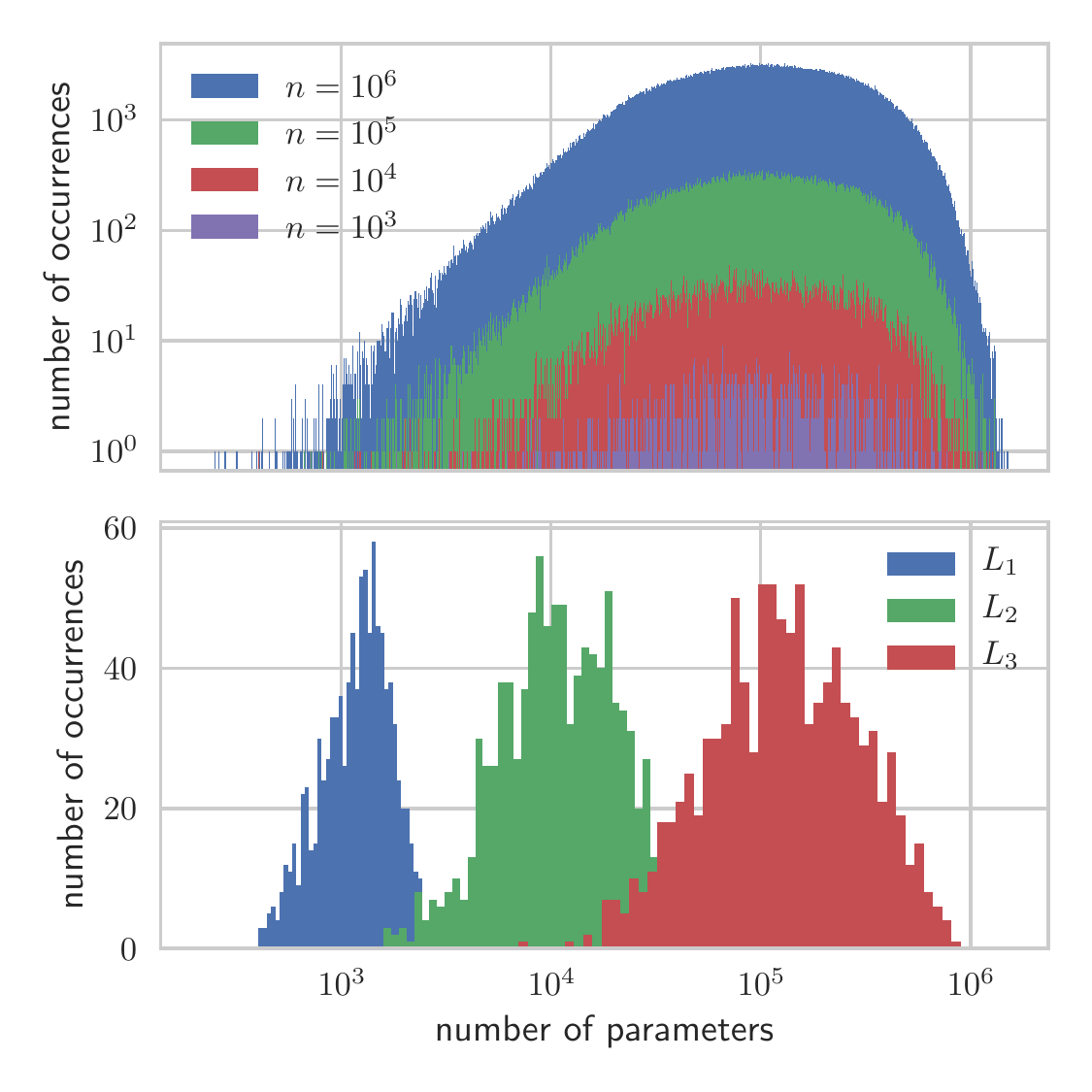}
	\caption{Statistics of number of parameters obtained when sampling up to one million networks from the base configuration space and when sampling 1000 networks from the restricted sampling laws.}
	\label{fig:fig_arch_serch_II}
\end{figure}

\section{Related work}
\label{sec:relatedwork}
Automated architecture search potentially discovers better models \cite{MIIKKULAINEN2019293,xie2017genetic,Zhong2017qlearn,Zoph16Reinforcement,Zoph_2018_CVPR_NASNet,cai2018efficient,BakerGNR16_dnn_reinforce,martin2019_arxiv}. 
However, traditional approaches require a vast amount of computing resources or cause excessive execution times due to full training of candidate networks\cite{real2017largescale}.
Early stopping based on learning curve predictors \cite{domhan2015_LCE} or transferring learned wights improves the timings \cite{wistuba2018deep}.
A method called train Train-less Accuracy Predictor for Architecture Search (TAPAS) demonstrates how to generalize architecture search results to new data without the need of training during the search process \cite{roiTAPAS_AAAI}. 
Architecture searches face the common challenge of defining the search space.
Historically, it happened that new networks are independently developed by expert knowledge that outperform previously found networks generated by architectural search.
In such cases, very expensive reconsiderations lead to follow up work to correctly account for a richer search space \cite{2018ENAS,2019CNAS}.
Recent progress in the field, such as MnasNet \cite{mnasnet_arxiv} and FBNet \cite{fbnet_arxiv} focus to tailor the search for smartphones by optimizing a multi-objective function including inference time. MnasNet trains a controller that adjusts to sample models that are more optimal according to the multi-objective. FBNet trains a supernet by a differentiable neural architecture search (DNAS) in a single step and claims to be $420\times$ faster since additional model training steps are avoided.   
In contrast to solving a joint optimization problem in one step, our proposed union of narrow-space searches follows a modular approach that separates the search process of finding architectures that strictly satisfy constraints from the training of candidate networks. That way, we can analyze ten-thousand architectures with zero training cost while only a small subset of suitable candidates are selected for training. 

Compression, quantization and pruning techniques reduce heavy computational needs based on the inherent error resilience of deep neural networks \cite{rybalkin2017hardware_blstm}.
Mobile nets \cite{mobilenets} or low-rank expansions \cite{Jaderber2014lowrank} change the topology into layers that require fewer weights and cause reduced workload.
Quantization studies the effect of using reduced precision floating point or fixed point formats \cite{Hill2016_rethinking,Loroch2017tensorquant}, compression further tries to reduce the binary footprint of activation and weight maps \cite{cavigelli2018bitplanecompression}, and pruning approaches avoid computation by enforcing sparsity \cite{Ashiquzzaman2019iot_nodepruning}. 
We use floatx, an IEEE 754 compliant reduced precision library \cite{floatx_submittedTOMS}, to assess data format specific aspects of networks.
The novelty of our work is that we jointly evaluate network topologies in combination with reduced precision.

\section{Core design procedures}
\label{sec:coredesign}

\subsection{Architecture search}

\begin{figure}[t]
	\centering
	\includegraphics[width=0.9\columnwidth]{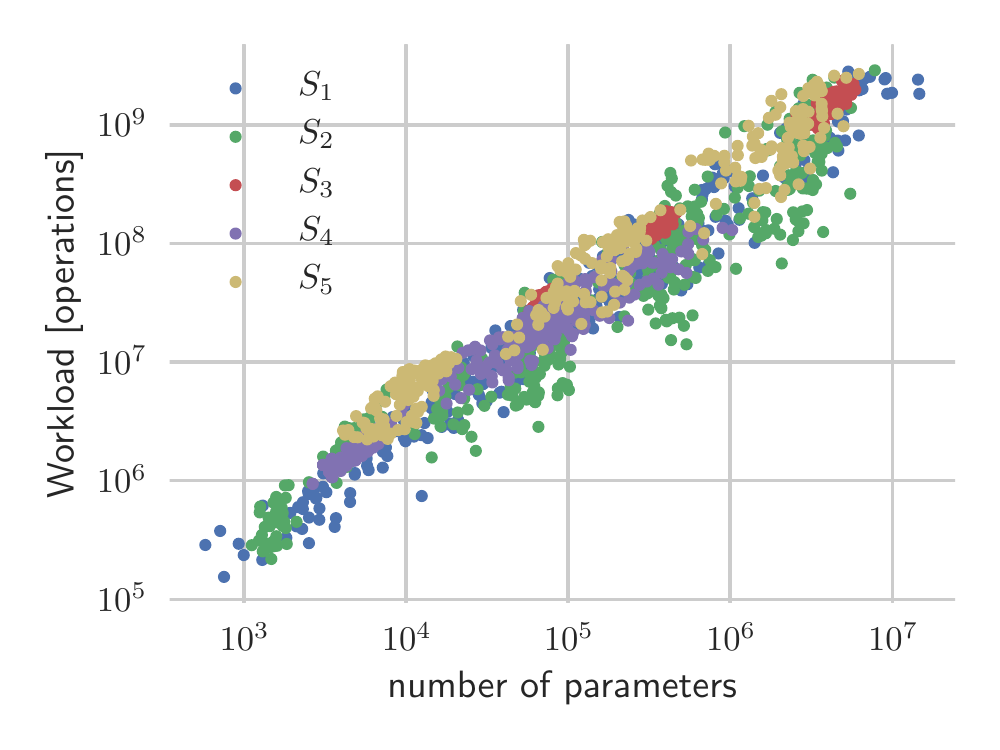}
	\caption{High correlations between the two analytical properties of network architectures.}
	\label{fig:fig_picalibration_a}
\end{figure}
\begin{figure}[t]
	\centering
	\includegraphics[width=0.9\columnwidth]{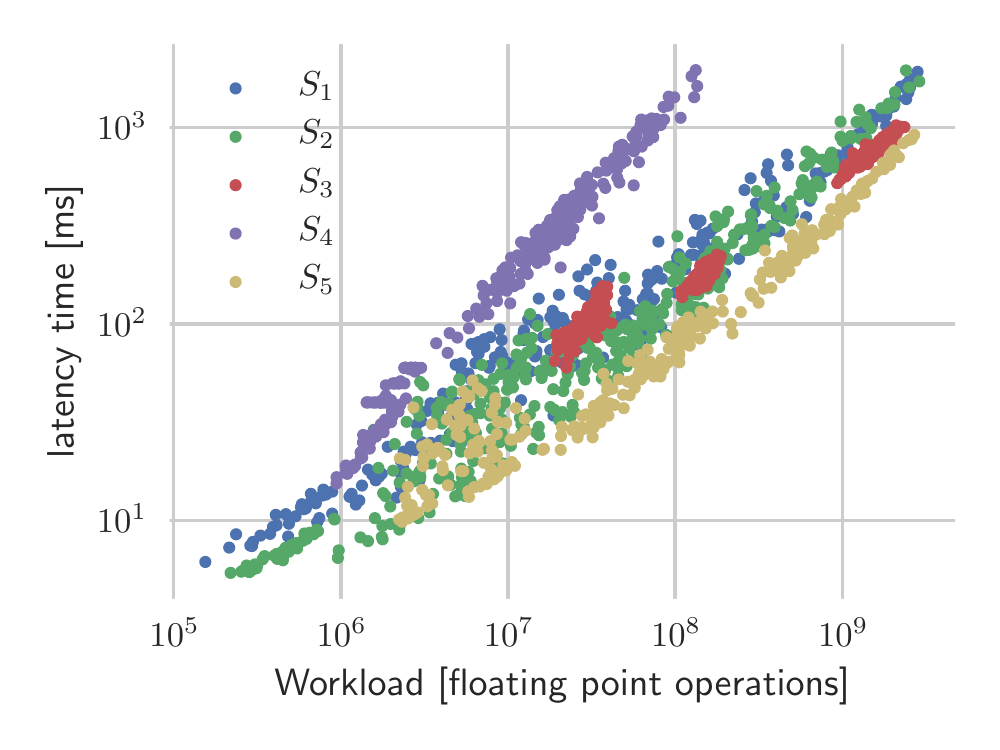}
	\caption{The run time dependent latency is best correlated with the workload where different search space specific characteristics are present.}
	\label{fig:fig_picalibration_c}
\end{figure}

It's challenging to define a space $S$ that produces enough variation and simultaneously reduces the probability of sampling suboptimal networks.
We propose narrow-space architecture searches, where results are obtained over aggregation of $n$ independent searches $S = \bigcup_{i=1}^{n} S_i$.
Since a good search space should satisfy $S_r \subset S$ where $S_r = \{M_1, ..., M_n\}$ is a set of reference models, 
we construct $S$ by designing narrow-spaces that obey  $M_i \in S_i$ in order to guarantee $S_r \subset S$.
Instead of considering superpositions, we have specialized search spaces that produce simple sequence structure with residual bypass operations (ResNets \cite{He_2016_resnet}) to even high fan-out and convergent structures such as they occur in the Inception module \cite{szegedy2015googlenet} or in DenseNets \cite{huang2017densely}.
The aggregation allows extending results easily with a tailored narrow-space search for new reference architectures.  
Next, we define a set of distribution law configurations $L_1(S_i), ..., L_k(S_i)$ that allow drawing samples in a biased way such that models satisfy properties of interest. 
Figure~\ref{fig:fig_arch_serch_I} demonstrates with an example the advantages over a uniform distribution among valid networks.
Consider a space of three-layer networks with allowed variations in kernel shapes in $\{1,3,5,7\}$ and output channels in $[1, 128]$ leading to $\card{S} = 4^6*128^3 = 8.6*10^9$ network configurations. 

Figure~\ref{fig:fig_arch_serch_II} shows the statistics over up to $10^6$ samples compared against sampling only 1000 samples when using restricted samplers $L_1, L_2$ and $L_3$.
The restricted random laws enable to efficiently generate networks of interest in contrast to the uniform sampler that fails to deliver high sampling densities in certain regions.
For example, only 132 out of $10^6$ networks have less than $1000$ parameters.

We define each narrow-space architecture search and its sampling laws according to the following design goals: first, the original model is included in the search space, second, only valid models are generated with a topology that resembles the original model, third, the main model-specific parameters are variated, fourth, the main way to generate small and efficient models was achieved through lowering channel widths in convolutional layers, and fifth, all random laws follow a uniform distribution over available options where the lower and upper limits where used as way to bias the models to span several orders of magnitude targeting the range of parameter and flop counts that are relevant for IoT applications. 

\subsection{Precision analysis}

The precision analysis evaluates model accuracies when models are running with reduced precision representations.
To follow a general methodology, we perform the precision analysis on the backend device that has different execution capabilities than current or future targeted IoT devices. 
The methodology enforces to use emulated computation throughout the analysis to assess accuracy independent of the target hardware.  
Low precision can be applied to model parameters, to the computations performed by the models and to the activation maps that are passed between operators. 
In this work, we follow the extrinsic quantization approach \cite{Loroch2017tensorquant}, where we enforce a precision caused by the reduced type $T_{w,t}$ of storage width $1 + w + t$ to be applied to all model parameters and all activation maps that are passed between operations.
For the analysis, we follow the  IEEE 754 standard \cite{2008ieee754} that defines storage encoding, special cases (Nan, Inf), and rounding behavior of floating-point data. 
A sign $s$, an exponent $e$ and the significand $m$ represent a number $v = (-1)^s * 2^e * m$ where the exponent field width $w$ and the trailing significant field width $t$ limit dynamic range and precision. 
Types $T_{5,10}$ and $T_{8,23}$ correspond to standard formats \emph{half} and \emph{float}.
Our experiments are based on a PyTorch \cite{pytorch_www} integration of the GPU quantization kernel based on the high performant floatx library \cite{floatx_submittedTOMS} that implements the type $T_{w,t}$.
The fast realization of the precision analysis allows elaborating over 3'000 models with a full grid search of 214 types ($w \in [1,8], t \in [1-23]$) on the full validation data.

\subsection{Deployment and performance characterization}

\begin{figure*}[t!]
	\centering
	\includegraphics[width=0.8\linewidth]{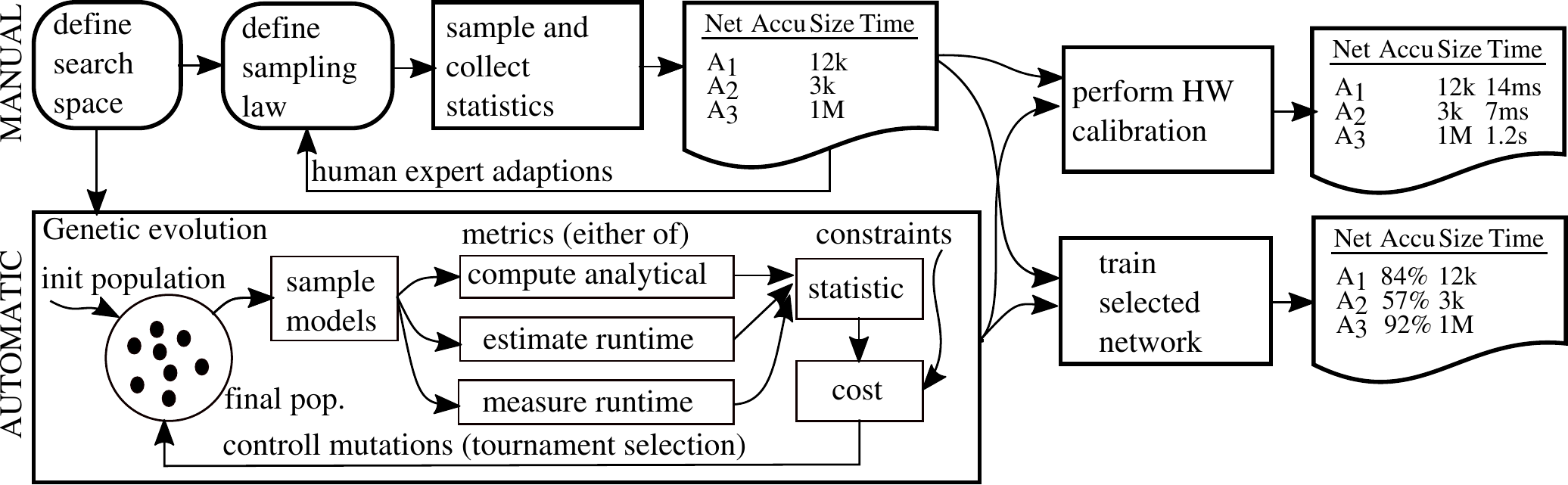}
	\caption{Manual and automatic workflow. First, sampling laws are defined to generate models of interest. Second, models are calibrated to check latency on the IoT device even if they are not yet trained. Third, models are trained to obtain their accuracy. Since training is the most expensive task, it is essential to reduce the amount of trained model to candidates of interest only.
	}
	\label{fig:fig_arch_manual_auto_pipes}
\end{figure*}

%
To evaluate model execution performance on the IoT target device we propose to perform a calibration to asses the execution speed of models of interest.
Despite many choices of deep learning frameworks, ways of optimizing code depending on compilation or version of software and even several hardware platforms that accelerate deep learning models, we formulate the performance characterization general and as most decoupled from the topology architecture search and the precision analysis to ease later extensions.
Performance measurements on the IoT device are affected by explicit and implicit settings. 
In this work, we demonstrate our search algorithm with performance measurements with the least amount of assumptions and requirements on the runtime.  
To that end, we selected a Raspberry-Pi 3(B+) as a representative IoT device. It features a Broadcom BCM2837B0, quad-core ARMv8 Cortex-A53 running at 1.4 GHz and the board is equipped with 1GB LPDDR2 memory \cite{pi3b_www}. The Raspberry-Pi 3(B+) belongs to the general-purpose device category that is shipped with peripherals (WiFi, LAN, Bluetooth, and USB, HDMI), a full operating system (Raspbian, a Linux distribution) available for a low cost of about $\$35$ per device \cite{mittal2019survey_jetson}.
Throughout this work, we measure the model inference latency on the target device by averaging over 10 repetitions.
We used a batch size of one to minimize latency and internal memory requirements. 
The latency study covers many relevant use cases, for example, the classification of sporadically arriving data in short time to prolong battery lifetime or frame processing of a video stream where the classification has to be completed before the next frame arrives.

For each model we consider two analytical properties, the number of trainable parameters and the workload measured as the number of floating-point operations required for inference.
The calibration relates analytical properties to execution performance and allows to separate runtime metrics. 
Figure~\ref{fig:fig_picalibration_a} and Figure~\ref{fig:fig_picalibration_c} show high correlations between the number of parameters, the workload and the measured latency on the Raspberry-Pi 3(B+). 
Workload and parameters follow a similar scaling over five orders of magnitude with homogenous variations.
The dynamic range of the latency spans more than two orders of magnitude with higher variations for larger models. 
However, due to the compute-bound nature of the kernels, the workload is the better latency time indicator than the number of parameters.

\begin{figure}[t] 
	\centering
	\includegraphics[width=0.9\columnwidth]{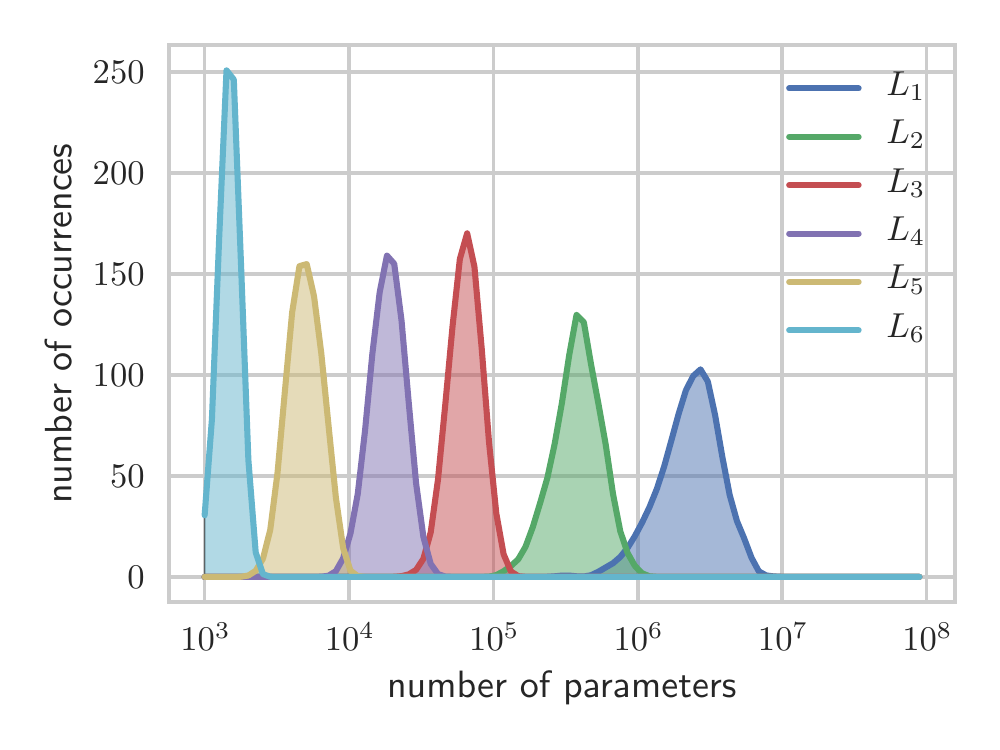}
	\caption{Manual defined sampling laws cover the full space.}
	\label{fig:fig_MobileNetV2_manual}
\end{figure}

\section{Fast cognitive design algorithms}
\label{sec:cognitive}

\begin{figure}[bh!]
	\centering
	\includegraphics[width=0.9\columnwidth]{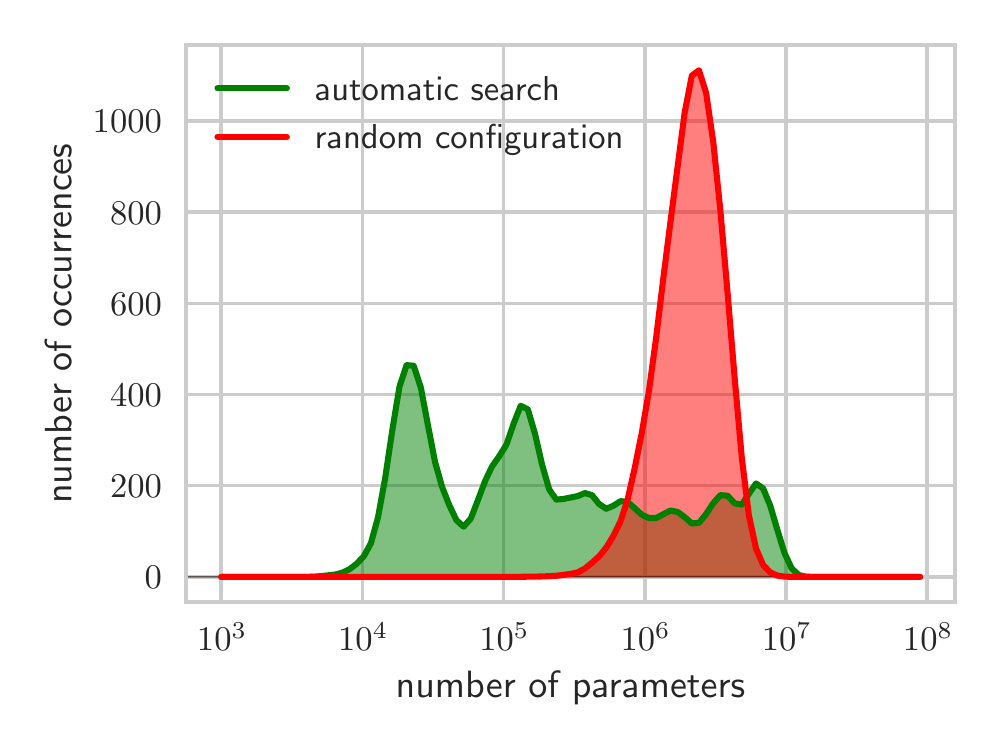}
	\caption{The automatic search finds configurations without human interaction and the distribution covers an higher dynamic range than just sampling uniformly.}
	\label{fig:fig_MobileNetV2_auto}
\end{figure}

In this section, we leverage the architecture search, the precision analysis, and the HW calibration to synthesize use case-specific solutions that satisfy given constraints. 
We address two tasks: First, the constraint search solves for the best model that satisfy given constraints. Second, the Pareto front elaboration provides insights into trade-offs over the full solution space. 
The two tasks are related. Solving the first task on a grid of constraints provides solutions to the second task while filtering the latter based on the given constraints allows returning to the former. 
Both tasks are solved in a manual and automated way as shown in Figure~\ref{fig:fig_arch_manual_auto_pipes}.
In the manual task, the expert user defines the narrow-space search and for each space a list of sampling laws. Collected statistics over analytical network properties provide quick feedback to adapt the settings to cover the range of interest. 
%

\begin{figure*}[t]
	\centering
	\includegraphics[width=0.7\linewidth]{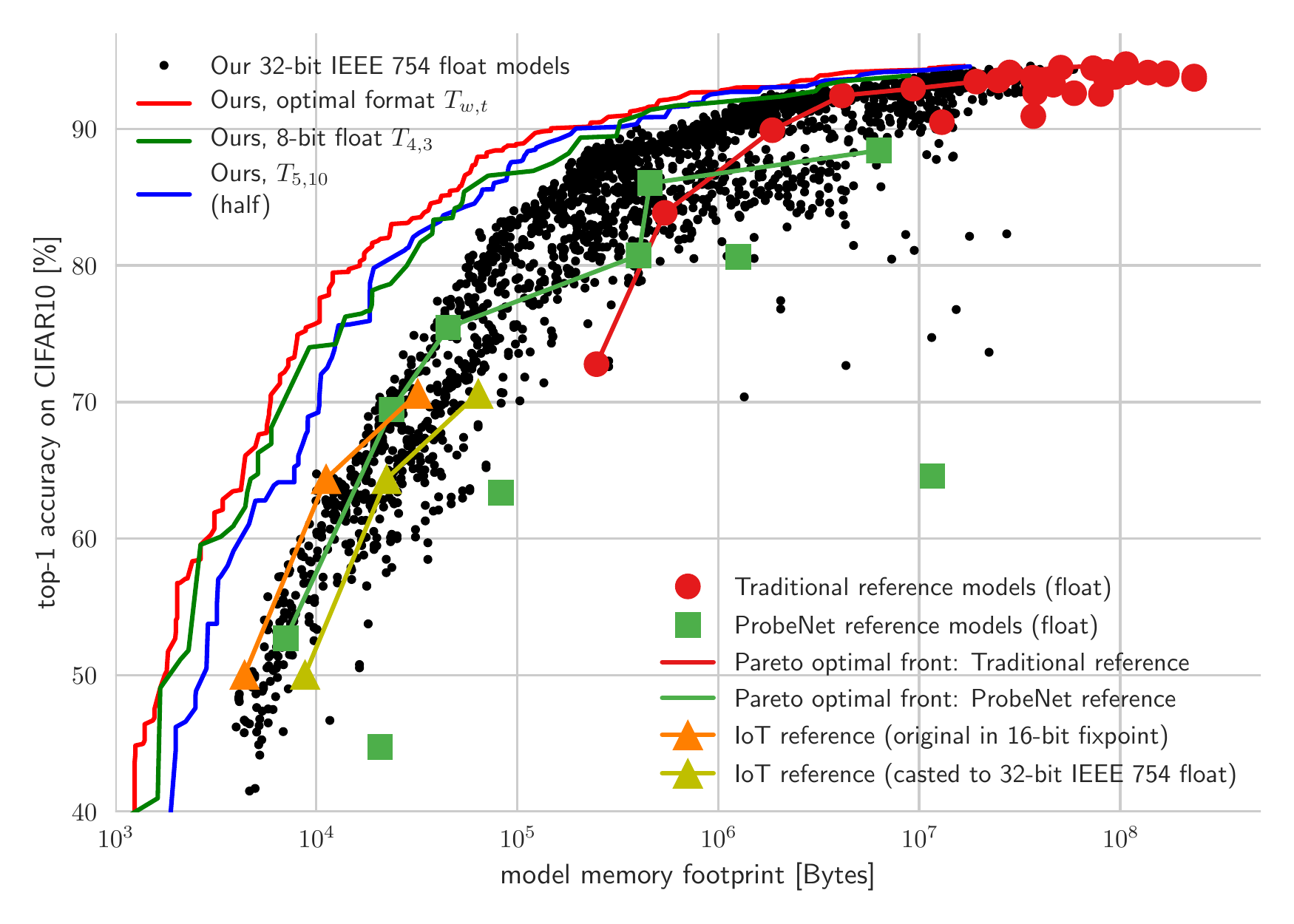}
	\caption{Results of our architecture search compared against reference models. Each dot represents a model by its size and the obtained accuracy on the CIFAR-10 validation set. Our search finds results over five order of magnitudes and especially finds various models that are much smaller than out-of-the box available models. In the restricted IoT domain, our search delivers models that outperform the reference with a wide margin for fixed constraints.}
	\label{fig:fig_result_tradeoff}
\end{figure*}

\begin{figure*}[th]
	\centering
	\includegraphics[width=0.7\linewidth]{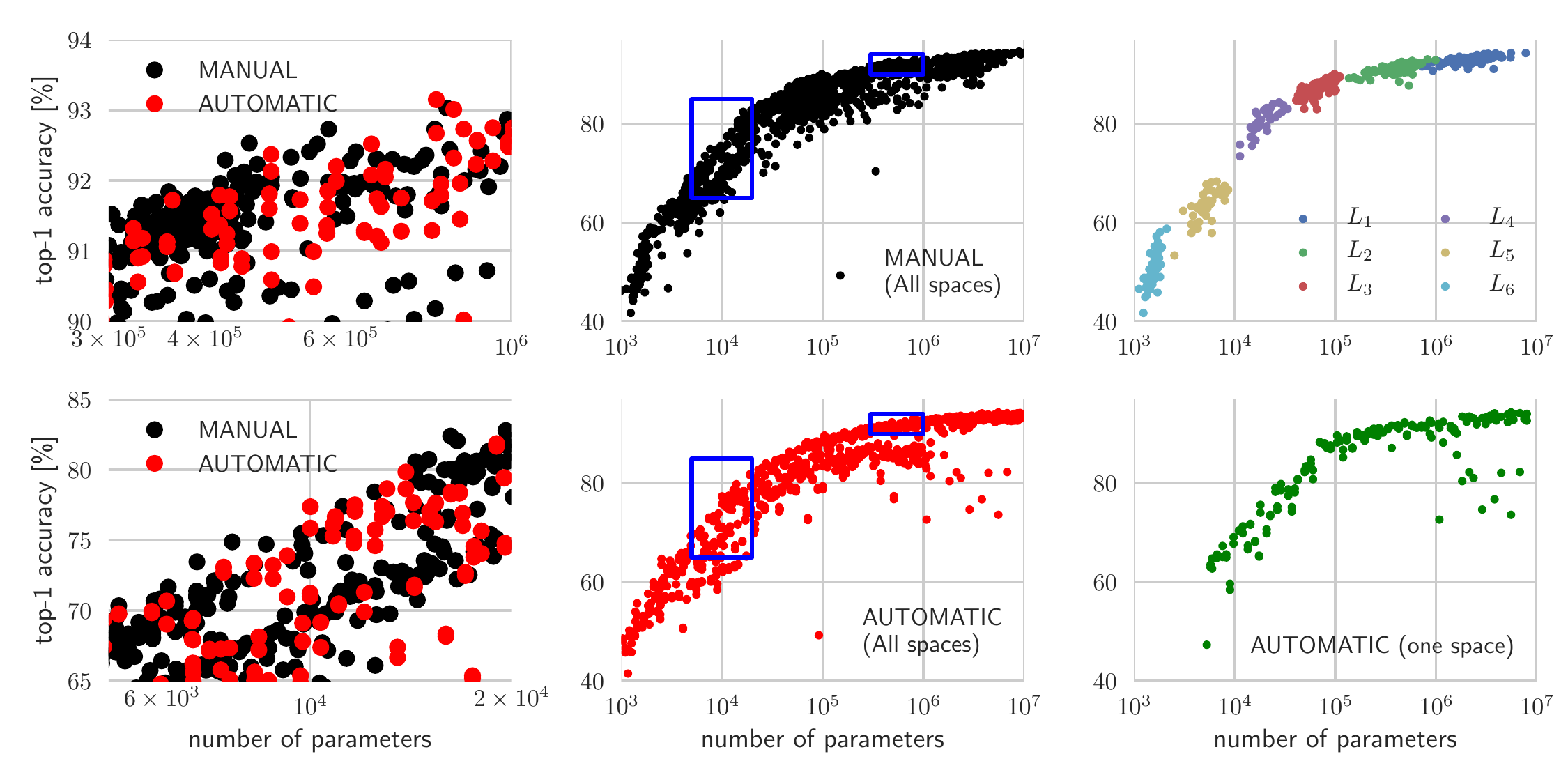}
	\caption{Left: zoomed view of direct comparison, manual and automatic search perform equally well. Middle: manual and automatic search results. In the manual case clusters or visible while the automatic search was able to sample more homogeneously. Right: results for one narrow-space search with marked clusters matching Figure~\ref{fig:fig_MobileNetV2_manual} and Figure~\ref{fig:fig_MobileNetV2_auto}.}
	\label{fig:fig_result_6plots}
\end{figure*}

Additionally, network run time metrics can be measured on the target device or estimated from calibration measurements. Next, depending on the task type, either a few candidate networks that satisfy constraints or a full wave of networks are selected for training. Large scale training takes the most time, each training job is of complexity $O(n_{train}C_{model}E)$, proportional to the amount of training data, the model complexity and the number of epochs the model is trained for. 

We designed a genetic and clustering based algorithm to automatize the design of sampling laws.
We define the valid space with a list of variables with absolute minimal and maximal ratings.
A sampling law $L(S_i)$ is defined as an ordered set of uniform sampling laws $L =  (U_x[l_x, h_x], ... )$ with lower and upper limits $l_x$ and $h_x$ per variable $x$. 
The genetic algorithm automatically learns the search space specific sampling law limits $[l_x, h_x]$. 
The cost function is defined in a two step approach. 
First, the statistic $(\mu_m, \sigma_m) := E^n_m(L)$ is estimated by computing means and standard deviations over the metric $m$ extracted from the $n$ generated topologies.
Second, cost is computed as  $c((\mu_m, \sigma_m), (\tau_1, \tau_2)) := \card{\mu_m - \sigma_m - \tau_1} + \card{\mu_m + \sigma_m - \tau_2}$ in order that the high density range of the estimated distribution coincides with a given interval $(\tau_1, \tau_2)$. 
We avoided definitions based on single sided constraints like $\mu < \tau$ since such formulations might be either satisfied trivially (using the smallest network) or satisfied by undesirable laws having wide or narrow variations.
We used the tournament selection variant of genetic algorithms \cite{goldberg1991_genalg} and defined mutations by randomly adapting the sampling law of hyper-parameters $l_x$ and $h_x$. We used an initial population of $n_{init} = 100$ and run the algorithm for $n_{steps} = 900$ steps while using $n_{eval} = 10$ samples to estimate mean and standard deviation per configuration. This way, one search considers $(n_{init} + n_{steps})*n_{eval} = 10'000$ networks. Since the final population might contain different sampling laws of similar quality, we perform spectral clustering \cite{stella2003spectralclustering} to find $k=10$ clusters with similar sampling laws. We assemble a list of the most different top-$k$ laws by taking the best fitted law per cluster. 

To elaborate the full search space with a Pareto optimal front, we split each decade into three intervals $[\tau, 2\tau, 5\tau, 10\tau]$ and define a grid for $\tau = 10^3, 10^4, 10^5, 10^6$ spanning five orders of magnitude.

We run the genetic search algorithm multiple times by setting the target bounds $(\tau_1, \tau_2)$ in a sliding window manner over consecutive values from the defined grid. Finally, we accumulate results from 12 genetic searches each found 10 sampling laws, where we sampled each law $n_{val}=100$ times to obtain the statistic of $12'000$ network architectures per narrow-space search. Figure~\ref{fig:fig_MobileNetV2_manual} and Figure~\ref{fig:fig_MobileNetV2_auto} show results for manual and automatic sampled networks. Even though the manual search allows to nicely cover the region of interest, human expertise is required to correctly define the parameters of the laws $L_1$ up to $L_6$. The naive sampling approach in the full search space produces a narrow distribution and is highly skewed towards larger networks. In contrast, the genetic algorithm was able to equalize the distribution and provides samples that cover much higher dynamic ranges, especially extending the scale for smaller networks without manually restricting the architecture.

\section{Results}
\label{sec:results}

\begin{figure}[ht]
	\centering
	\includegraphics[width=0.9\columnwidth]{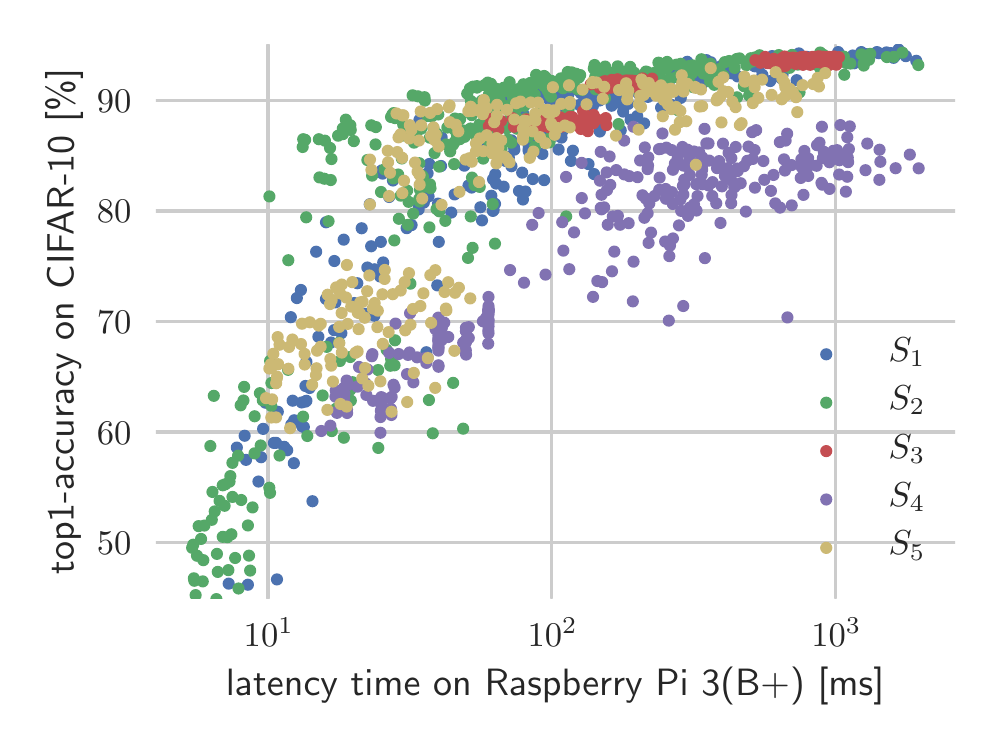}
	\caption{Final result showing the achievable tradeoffs between on the IoT device measured model latency and the model accuracy. Our search is able to deliver models that run below 10ms on the Raspberry Pi 3(B+) which we consider as representative cost limited IoT device.}
	\label{fig:fig_result_pi_manual}
\end{figure}

\begin{figure*}[th]
	\centering
	\includegraphics[width=.8\linewidth]{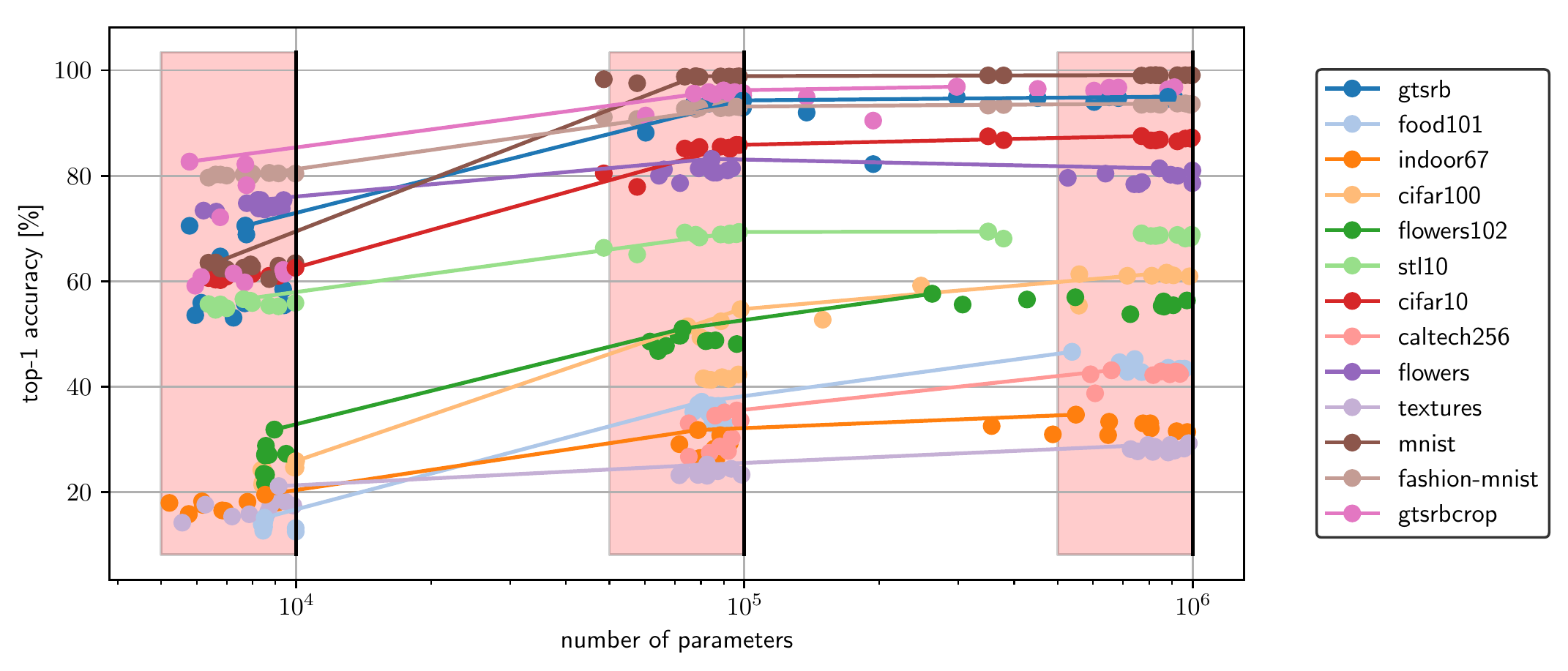}
	\caption{We demonstrate scalability of our approach by applying our search to three constraints on 13 datasets. Best models per dataset and constraint are connected with a line.}
	\label{fig:fig_result_manydsets}
\end{figure*}

To study our algorithm we run full design space explorations on the well established CIFAR-10 \cite{data_cifar10_100} classification task and compare our results with those obtained with established reference models. Figure~\ref{fig:fig_result_tradeoff} shows the trade-off between the model size and the obtained accuracy including manual and automatic generated results of the aggregate search spaces. The Pareto optimal front follows a smooth curve that saturates towards the best accuracy obtainable for large models. The number of parameters is logarithmic and the accuracy linearly scaled. Even very small models with less than 1000 parameters can achieve above $45\%$ of accuracy. The accuracy increase per decade of added parameters is in the order of $30\%$, $15\%$, $3\%$ and $<2\%$ points and diminishes very quickly. This effect allows constructing models that consist of multiple orders of magnitude fewer parameters and provides economical interesting solutions when IoT devices are powerful enough to process data in real-time. We compare our results with three sources of reference models, a) with traditional reference models, b) with ProbeNets \cite{dcn_submitted_ComputerVision} that are designed to be small and fast and, c) with models that were designed and run on the parallel ultra-low power (PULP) platform \cite{Conti2016PULP}. Traditional models include 30 reference topologies including variants of VGG \cite{vgg_arxiv}, ResNets \cite{he2016deep_resnet}, GoogleNet \cite{Szegedy_2016_CVPR_inception}, MobileNets \cite{MobileNets_arxiv} dual path nets (DPNs) \cite{DPN2017_nips} and DenseNets \cite{Huang_DenseNets_2017_CVPR} where most of them (28/30) exceed 1M parameters. ProbeNets are originally introduced to characterize the classification difficulty and are by design considerably smaller \cite{dcn_submitted_ComputerVision}. They act as reference points for manual designed networks that cover the relevant lower tail in terms of parameters. In the IoT relevant domain (<10M parameters) our search outperforms all the listed reference models. 
The top three fronts in Figure~\ref{fig:fig_result_tradeoff} show the results of the precision analysis. For each trained model we evaluated the effect of running models with all configurations of type $T_{w,t}$ and we extract and plot the Pareto-optimal front. We considered three cases, running all models with half precision, running all models with the type $T_{4_3}$ which is the best choice for types of 8-bit length and running each model with its individual best trade-off type $T_{w,t}$.
We empirically demonstrate that running with reduced precision pushes the Pareto optimal front. Under a given memory constraint, accuracy improves by over $7\%$ points for half and over $1\%$ points further for running with the model individual format.  
Figure~\ref{fig:fig_result_6plots} shows details about manual and automatic searches both leading to very similar results. The right figure shows results obtained for one narrow-space search, where manually defined sampling laws lead to clusters. The automatic search was able to homogeneously cover a similar range. Figure~\ref{fig:fig_result_pi_manual} shows inference times when the same set of models is executed on the Raspberry Pi 3(B+). Similarly, towards the small model end of the scale, given additional time for the latency results in dominant accuracy gains, however towards the traditional high accuracy domain, even slight accuracy improvements are only achieved with even more complex models that cause long evaluation times.   
Figure~\ref{fig:fig_result_manydsets} demonstrates the scalability of our approach. We applied our search for three constraints $\tau = 10^3, 10^4, 10^5$ on thirteen datasets \cite{dcn_submitted_ComputerVision} where we spend a training effort of ten architectures per dataset and constraint. The lines connect the best per constraint and dataset performing architectures.  
\section{Conclusion}
\label{sec:conclusion}
We studied the solution of synthesizing deep neural networks that are eligible candidates to efficiently run on IoT devices.
We propose a narrow-space search approach to quickly leverage knowledge from existing architectures that is modular enough to be further adapted to new design patterns.
Manually and automatically designed sampling laws allows generating various models with the number of parameters covering multiple orders of magnitude.
We demonstrate that reduced precision improves top1 accuracy by over $8\%$ points for constraint weight memory in the IoT relevant domain.
A strong correlation between model size and latency enables to create small enough models that provide superior inference response latencies below $10\mathrm{ms}$ on a $\$35$ edge device.     

\subsubsection*{Acknowledgments}
This work was funded by the the European Union’s H2020 research and innovation programme under grant agreement No 732631, project OPRECOMP.

\fontsize{9pt}{10pt} \selectfont 	

\bibliography{main.bib}   
\bibliographystyle{aaai}  

\end{document}